\pdfoutput=1

\documentclass[11pt]{article}

\usepackage[preprint]{acl}

\usepackage{times}
\usepackage{latexsym}
\usepackage{amssymb}
\usepackage{utfsym}

\usepackage[T1]{fontenc}

\usepackage[utf8]{inputenc}

\usepackage{microtype}

\usepackage{inconsolata}

\usepackage{graphicx}
\usepackage{soul}
\usepackage{url}
\usepackage{hyperref}
\usepackage{caption}
\usepackage{amsmath}
\usepackage{array} 
\usepackage{amsthm}
\usepackage{booktabs}
\usepackage{algorithm}
\usepackage{algorithmic}
\usepackage{multirow}

\usepackage{algorithmic}
\usepackage[switch]{lineno}
\usepackage{color, colortbl}
\definecolor{Gray}{gray}{0.9}
\usepackage{tcolorbox}
\tcbuselibrary{breakable}
\usepackage{soul}
\usepackage{makecell}
\usepackage{listings}

\floatstyle{ruled}
\newfloat{listing}{tb}{lst}{}
\floatname{listing}{Listing}

\usepackage{lipsum}

\usepackage{xcolor}
\usepackage{xspace}
\usepackage{multirow}
\usepackage{multicol}

\usepackage[capitalize]{cleveref}

\def\modelname{MMDuet\xspace}
\def\datasetname{MMDuetIT\xspace}

\title{VideoLLM Knows When to Speak: Enhancing Time-Sensitive Video Comprehension with Video-Text Duet Interaction Format}

\author{
 \textbf{Yueqian Wang\textsuperscript{1}},
 \textbf{Xiaojun Meng\textsuperscript{2}},
 \textbf{Yuxuan Wang\textsuperscript{3}},
 \textbf{Jianxin Liang\textsuperscript{1}},
\\
 \textbf{Jiansheng Wei\textsuperscript{2}},
 \textbf{Huishuai Zhang\textsuperscript{1,4}},
 \textbf{Dongyan Zhao\textsuperscript{1,4}},
\\
 \textsuperscript{1}Wangxuan Institute of Computer Technology, Peking University \\
 \textsuperscript{2}Huawei Noah’s Ark Lab
 \textsuperscript{3}Beijing Institute for General Artificial Intelligence \\
 \textsuperscript{4}State Key Laboratory of General Artificial Intelligence \\
\\
 \small{
   \textbf{Correspondence:} \href{mailto:zhanghuishuai@pku.edu.cn}{zhanghuishuai@pku.edu.cn}, \href{mailto:zhaodongyan@pku.edu.cn}{zhaodongyan@pku.edu.cn}
 }
}

\usepackage{xcolor}

\begin{document}
\maketitle
\begin{abstract}
Recent researches on video large language models (VideoLLM) predominantly focus on model architectures and training datasets, leaving the interaction format between the user and the model under-explored. In existing works, users often interact with VideoLLMs by using the entire video and a query as input, after which the model generates a response. This interaction format constrains the application of VideoLLMs in scenarios such as live-streaming comprehension where videos do not end and responses are required in a real-time manner, and also results in unsatisfactory performance on time-sensitive tasks that requires localizing video segments.
In this paper, we focus on a video-text duet interaction format. This interaction format is characterized by the continuous playback of the video, and both the user and the model can insert their text messages at any position during the video playback. When a text message ends, the video continues to play, akin to the alternative of two performers in a duet.
We construct \datasetname, a video-text training dataset designed to adapt VideoLLMs to video-text duet interaction format. We also introduce the Multi-Answer Grounded Video Question Answering (MAGQA) task to benchmark the real-time response ability of VideoLLMs.
Trained on \datasetname, \modelname demonstrates that adopting the video-text duet interaction format enables the model to achieve significant improvements in various time-sensitive tasks (76\% CIDEr on YouCook2 dense video captioning, 90\% mAP on QVHighlights highlight detection and 25\% R@0.5 on Charades-STA temporal video grounding) with minimal training efforts, and also enable VideoLLMs to reply in a real-time manner as the video plays.
\end{abstract}

\section{Introduction} \label{sec:intro}

\begin{figure*}
    \centering
    \includegraphics[width=\linewidth]{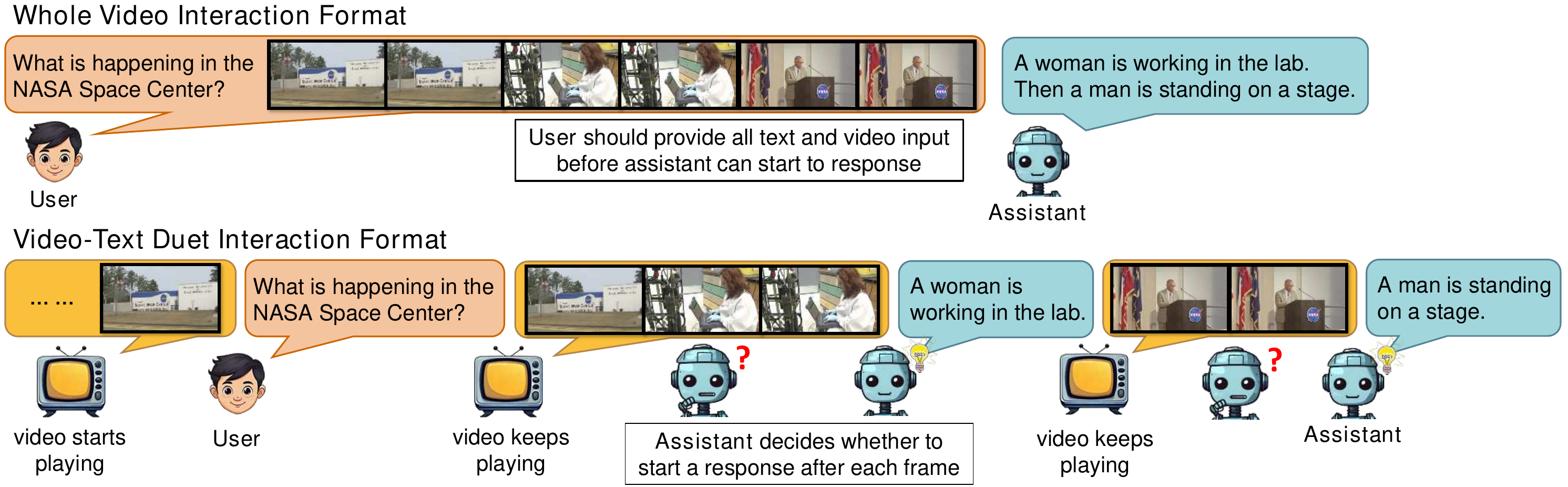}
    \caption{An example of the common \textit{Whole Video} Interaction Format and our Video-Text Duet Interaction Format.}
    \label{fig:format}
\end{figure*}

Videos are becoming an increasingly important medium to acquire information on a daily basis. 
Powered by recent advancements in large language models (LLMs) \cite{Touvron2023Llama2O,Jiang2023Mistral7,Shao2024DeepSeekV2AS,Dubey2024Llama3,Yang2024Qwen2TR} and vision encoders \cite{Radford2021LearningTV,Zhai2023SigmoidLF,Sun2023EVACLIPIT,Oquab2023DINOv2LR,Wang2024InternVideo2SV}, several video large language models (VideoLLM) \cite{Li2023MVBenchAC,Liu2024STLLMLL,Li2024LLaVANeXTInterleaveTM,Li2024LLaVAOneVisionEV,Zhang2024LLaVAVideo,Wang2024EfficientTE} have already demonstrated strong abilities for holding conversations and answering questions about videos. 
A common feature of these models is using visual encoders to encode all frames sampled from the entire video at first, and integrate them into text input by concatenating them to input embeddings or using cross attention.

Recent research on VideoLLMs has primarily concentrated on model architectures and training datasets, with limited exploration of the interaction format between the user and the model. In this paper, the ``interaction format'' of VideoLLMs comprises the following two aspects: (1) a \textbf{chat template} used to convert input sources, e.g., video, user text query, and model response, into a sequence of tokens; (2) a \textbf{turn-taking rule} organizing inputs of different sources to finalize an interaction format. For example, for most existing VideoLLMs, the interaction format is: (1) for the chat template, the model uses (frames sampled from) the full video and a text query as input, and then outputs
a response; (2) for the turn-taking rule, usually the model is permitted to take its turn to generate a response when both the whole video content and user query have ended, e.g., when an \textless{}eos\textgreater{} token is explicitly provided. We refer to this traditional interaction method as ``\textit{whole video}'' in the rest of this paper. 

However, this all-along used \textit{whole video} interaction has the following two defects, which hinder the performance and real-world usage scenarios of VideoLLMs:
Firstly, it does not admit timely interactions. As the video is often input as a whole, this limits its usage in more scenarios like live broadcasts or surveillance videos, in which the video does not end at a specific time. Even if we can segment the video into multiple fixed-length clips for input, the model still cannot generate responses in a real-time manner when necessary, as it does not know whether it is feasible and appropriate to reply at the end of this clip.
Secondly, it performs unfavorably on time-sensitive video comprehension tasks. In this paper we use \textbf{``time-sensitive tasks''} to refer to tasks in which the model is required to provide responses that include specific times in the video, such as temporal video grounding \cite{Krishna2017DenseCaptioningEI,Gao2017TALLTA,Hendricks2017LocalizingMI}, video highlight detection \cite{Lei2021QVHighlightsDM}, dense video captioning \cite{Zhou2017TowardsAL,Krishna2017DenseCaptioningEI}, grounded video question answering \cite{Xiao2023NextGQA}, etc.

In this work, we formalize the Video-Text Duet Interaction Format, an interaction method that aims to enhance VideoLLMs by addressing the aforementioned issues. An illustration of the \textit{whole video} interaction format and the video-text duet interaction format is shown in \cref{fig:format}. With our video-text duet interaction format, the video is continuously played and input to the model frame-by-frame. Both the user and model can insert their text messages right after any frame during the video play.
When a dialogue turn from either the user or the model ends, the video stream can have the floor and input video frames to the model until another turn is started by either the user or the model, akin to the show of two performers in a duet.
This improves the timeliness of interaction and better suits real-world applications such as live-streaming or surveillance video comprehension. Moreover, by inserting responses to the video where is most relevant, the model can learn to generate responses by referencing a smaller but fine-grained fraction of the video before this position. In this manner, it facilitates information retrieval to describe lengthy videos, as well as enables a response to be ``grounded'' at the targeted position of the video. We believe this design contributes to addressing the above discussed issues of existing \textit{whole video} VideoLLMs.

To prove the effectiveness of the video-text duet interaction format, we construct \datasetname, a dataset to facilitate the training of a versatile VideoLLM following the video-text duet interaction format. We propose Multi-Answer Video Grounded QA (MAGQA), a novel task that requires the model to generate answers at appropriate timespans in a real-time manner to align with potential applications of live-streaming video comprehension.
We also train \modelname, a VideoLLM that implements our proposed video-text duet interaction format. Initialized with LLaVA-OneVision \cite{Li2024LLaVAOneVisionEV} and trained with \datasetname at a low cost, \modelname achieves significant performance improvement in various time-sensitive tasks, and is able to generate responses in real-time as the video plays.

\section{Related Works}

The advancement of large language models (LLMs) and visual encoders has led to numerous efforts on their integration, aiming to utilize the powerful understanding and generation abilities of existing LLMs for video-related tasks \cite{Li2023MVBenchAC,Liu2024STLLMLL,Li2024LLaVANeXTInterleaveTM,Li2024LLaVAOneVisionEV,Wang2024EfficientTE, ijcai2023p582}.
These models exhibit a decent ability of video understanding such as captioning or summarizing \cite{ijcai2023p582}. However, their performance on time-sensitive tasks is still unsatisfactory.


Recent works attempt to empower VideoLLMs with the ability to localize and represent segments in videos, and thus achieve better performance on tasks like temporal video grounding or dense video captioning. These works explore new ways on how to easily represent video clips with texts, such as second numbers of timestamp (TimeChat \cite{Ren2023TimeChatAT}), timeline percentage (VTimeLLM \cite{Huang2023VTimeLLMEL}) or using special textual tokens (VTG-LLM \cite{Guo2024VTGLLMIT}, Grounded-VideoLLM \cite{Wang2024GroundedVideoLLMSF}). However, their performance has not been satisfactory yet, possibly due to LLMs' limited ability to accurately count and generate numbers \cite{Schwartz2024NumeroLogicNE} to localize each video frame. To alleviate this issue, HawkEye \cite{Wang2024HawkEyeTV} uses a coarse-grained method by referring to a larger fraction of the video, but it requires multiple rounds of recursive grounding to precisely locate a segment and may not express multiple segments at a time.

The work most similar to our motivation is VideoLLM-Online \cite{Chen2024VideoLLMonlineOV}, which proposes a framework named LIVE for training VideoLLMs to interrupt video streams and insert responses. However, they only finetune a model on Ego4D \cite{Grauman2021Ego4DAT} and COIN \cite{Tang2019COIN} to demonstrate the LIVE training and inference, and do not explore on how the model capabilities vary with this new type of interaction, especially the zero-shot performance on time-sensitive tasks.

Our work differs from VideoLLM-Online at: Firstly, providing a more general description of the video-text dual interaction format, including a wider variety of criteria for determining whether a response should be generated, and its application on new tasks such as temporal video grounding and grounded question answering; Secondly, introducing a new dataset \datasetname and the method on building such datasets; Thirdly, proposing a new task MAGQA; Lastly, proposing a more powerful model \modelname that has state-of-the-art performance on various time-sensitive tasks and zero-shot generalization ability.

\section{The Video-Text Duet Interaction Format} \label{sec:format}

In \cref{sec:intro}, we have defined the concept of ``interaction format'' with two aspects (\textit{i.e.,} chat template \& turn-taking rule), as well as the drawbacks of the commonly-used \textit{whole video} interaction format. Now we re-emphasize and formalize our video-text duet interaction format, which is completely different from previous to implement VideoLLMs.

(1) For the chat template, inspired by but different from the LIVE framework which is used to implement VideoLLM-Online \cite{Chen2024VideoLLMonlineOV}, we consider the video stream as a conversation participant just like the role of user/assistant, and the input sequence consists of alternating turns among these three roles.
(2) For the turn-taking rule, when the turn of the user or assistant ends, the video stream can take the floor and start its turn to input video frames. When each single frame is consumed, both the user and the assistant role can interrupt the video stream at any time, and start its own turn to query or generate a response, as totally decided by the user or the assistant, respectively.



\section{\modelname: Our Proposed VideoLLM}
\subsection{Model Structure}
We propose \modelname, a model trained following the video-text duet interaction format, which can thus autonomously decide at what position in the video to generate what response.
Like almost all existing VideoLLMs, \modelname consists of three components: 1) a visual encoder that encodes sampled frames from the video to visual feature, 2) a linear projector that transforms the encoded visual feature to a list of visual tokens that is aligned into the LLM textual embedding space, and 3) a transformer-decoder-based LLM that takes both textual and visual tokens as input and uses its language modeling  head to predict the next token. 

The only difference in model structure between our \modelname and existing VideoLLMs is that we add two more heads in addition to the language modeling head (LM Head) of the LLM, namely the \textbf{informative head} and the \textbf{relevance head}, for determining whether to start a response after each frame. Each head is a linear layer and has a weight with shape $h \times 2$, where $h$ is the hidden size of the used LLM. Each head takes the final layer hidden state of the last visual token of each frame as input, and performs a binary classification.
To be specific, 1) the informative head is designed to predict how much new information is acquired upon viewing the current frame. If the model can obtain a ``significant amount'' of new information upon viewing a new frame (which we will further discuss in \cref{sec:data_dense_captioning}), it should classify this frame as \textit{TRUE} category; otherwise, it should classify it as \textit{FALSE}.
2) The relevance head is designed to predict whether the current frame is related to the user query. Similarly, \textit{TRUE} category means to be related, while \textit{FALSE} means not. We denote the probability of \textit{TRUE} category of informative head and relevance head as \textbf{informative score} and \textbf{relevance score} for each sampled video frame. These two scores will be used to decide whether the model (\textit{i.e.,} assistant role) should interrupt the video and start its own turn. Compared with VideoLLM-Online \cite{Chen2024VideoLLMonlineOV} that makes this decision by predicting one special token using the LM Head, our design has the following merits:
(1) The ground truth labels of informative scores and related scores are acquired based on the characteristic of the video itself, rather than on ad-hoc response decisions. Therefore, there are better labels for models to converge during training.
(2) By combining two scores we can flexibly set different criteria for response generation, rather than only relying on the logits of one special token; (3) The relevance head can be used to precisely perform temporal video grounding and highlight detection tasks, expanding the application scenarios of \modelname. 

\subsection{Inference Procedure} \label{sec:inference}
When consuming every single sampled frame of the video, we first check if there is a user query happening at this time. If yes, we first input this user turn to the model. Then the sampled frame is input to the model, after which the informative score and relevance score are calculated. We use a function \texttt{need\_response} to estimate whether the model should generate an assistant response according to the informative scores and relevance scores for this frame along with previous frames. If yes, the \texttt{generate} function of the LLM outputs a response. Different \texttt{need\_response} functions can be designed depending on the specific task, which is introduced in the experiment section (\cref{sec:experiments}). This process can be efficiently implemented by updating the KV Cache each time when a frame or text is input or generated, and a python-style sudo code is provided in \cref{sec:pesudo_code_inference}.



\section{\datasetname: Dataset for Training \modelname}
We build \datasetname, a dataset for training the \modelname model to learn to calculate the informative and relevance scores, and autonomously output replies at any necessary time in the play of the video. \datasetname is composed of three different types of tasks that benefit our model training: dense captioning, multi-answer grounded video question answering, and temporal video grounding. An example of the input format for each task is listed in \cref{sec:example_input}.

\subsection{Dense Captioning} \label{sec:data_dense_captioning}

\begin{figure}
    \centering
    \includegraphics[width=\linewidth]{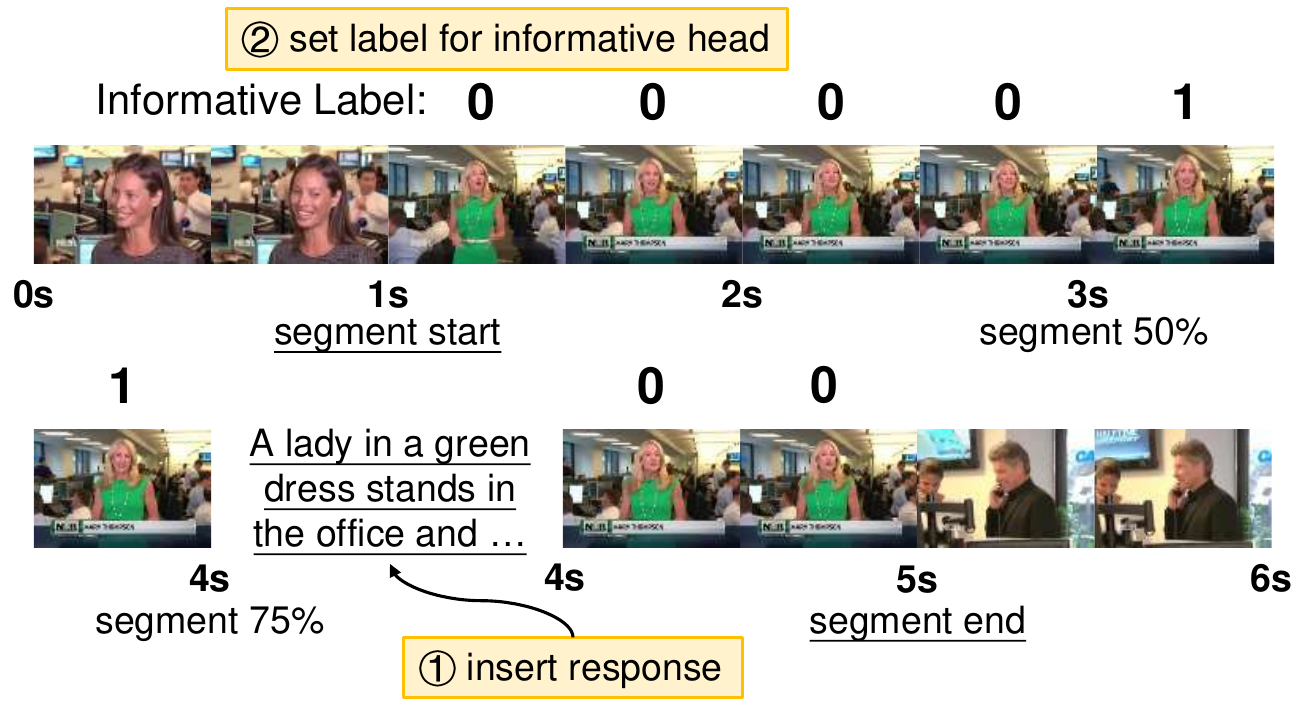}
    \caption{Example of reformatting the annotation of a video segment to video-text duet interaction format in \datasetname. Information from the original annotation is emphasized with underlines.}
    \label{fig:reformat_data}
\end{figure}

We use Shot2Story \cite{Han2023Shot2Story20KAN}, a video-text dataset with segment-level captions, as our dense captioning training data. Specifically, we use the 43k human-annotated subset due to its high-quality and detailed annotations.
We preprocess the data to serve our purposes, and an illustration of reformatting the video segment and caption annotations to video-text duet interaction format is in \cref{fig:reformat_data}: we randomly sample a position from 50\% to 75\% time duration for the corresponding video segment, and insert the caption at that position as a model response. 
We also create labels for the informative head in dense captioning tasks by setting the informative head’s label to \textit{TRUE} for frames between 50\% of this segment and the insertion point of the response, and set labels to \textit{FALSE} for the other frames.
To adapt to long video input, we also select videos with 2 to 4 minutes in length from COIN \cite{Tang2019COIN} as a dense captioning task to \datasetname. The annotations in COIN are reformatted using the same method as Shot2Story. 
For more details about this data reformat process please refer to \cref{sec:detail_reformat_process}.

\subsection{Multi-Answer Grounded Video QA} \label{sec:data_magqa}
An important application scenario for the video-text duet interaction format is multi-answer grounded video question-answering (MAGQA). Consider when we are watching a live broadcast of a basketball game and want to track the actions of a particular player in the game. This exemplifies a MAGQA task: the question is "What does this particular player do in the video?". Each time this player performs an action, the model should respond with a description of this action (\textit{i.e.,} multiple answers) in a real time manner. We believe this newly proposed MAGQA task can be widely used in real-world scenarios when users interact with a live-streaming video.

We construct training data for this task using GPT4o-2024-08-06 \cite{openai2024gpt4o}.
Given the captions of all segments from the video as input, GPT4o is prompted to generate a question related to one or more captions. For each of the segment captions, if it is related to the question, then GPT4o should also generate an answer that can be inferred from this caption. Otherwise, GPT4o should reply with ``\texttt{Not Mentioned.}'', and this answer is not added to the training data.
We use the same insertion method of dense captioning task as described in \cref{sec:data_dense_captioning}, to insert the answers into the video stream and construct informative head labels, and the question is inserted at a random place before the first answer.
We also use the same insertion method to convert the human-annotated Shot2Story test set and randomly sampled 2000 examples as the test set of our MAGQA benchmark in \cref{sec:exp_magqa}. Therefore, this dataset contains 36834 examples in the train set and 2000 examples in the test set, and we name it as ``Shot2Story-MAGQA-39k''. 

We have manually checked its data quality, and details of this process are stated in \cref{sec:data_quality}.

\subsection{Temporal Video Grounding}
We also add DiDeMo \cite{Hendricks2017LocalizingMI}, $\text{HiREST}_{grounding}$ \cite{Zala2023HierarchicalVR} and QuerYD \cite{Oncescu2021QUERYDAV}, three temporal video grounding tasks in \datasetname. Note that these data are used only for training the relevance head, which is designed for performing temporal video grounding tasks and judging the relevance between the question and the video for QA tasks. The query is first added at the beginning of the input sequence. For frames that are annotated as relevant to the query, we set the relevance head’s label to \textit{TRUE}; otherwise, we set it to \textit{FALSE}. 

\subsection{Dataset Statistics}

\begin{figure*}[h!]
\centering
\begin{minipage}{0.27\textwidth} 
    \centering
    \includegraphics[width=\linewidth]{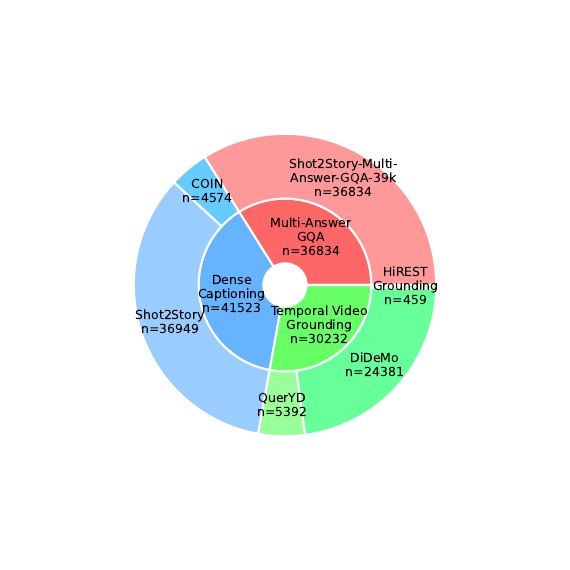}
    \caption{Data Distribution of \datasetname.}
    \label{fig:data_distribution}
\end{minipage}%
\hfill 
\begin{minipage}{0.68\textwidth}
    \centering
    \small
    \begin{tabular}{c|c|c||c}
        \toprule
         & QVHighlights & Charades-STA & YouCook2 \\
         & mAP/HIT@1 & R@IoU=0.5/0.7 & SODAc/CIDEr/F1 \\
        \hline
        Video-LLaMA & 11.3/15.6 & 2.7/1.2 & 0.0/0.0/0.1 \\
        VideoChat-Embed & 13.1/18.1 & 3.2/1.4 & 0.2/0.6/3.4 \\
        VideoChatGPT & - & 7.7/1.7 & - \\
        \hline
        TimeChat & 14.5/23.9 & 32.2/13.4 & 1.2/3.4/12.6 \\
        VTimeLLM & - & 31.2/11.4 & - \\
        HawkEye & - & 31.4/14.5 & - \\
        VTG-LLM & 16.5/33.5 & 33.8/15.7 & 1.5/5.0/17.5 \\
        VideoLLM-Online & - & - & 0.4/0.9/5.8 \\
        \hline
        LLaVA-OV-TC & 17.6/32.9 & 33.1/12.4 & 1.9/3.3/\textbf{21.8} \\
        LLaVA-OV-VT & 19.0/40.0 & 36.5/12.3 & 2.5/6.7/\textbf{14.0} \\
        \hline
        \modelname(Ours) & \textbf{31.3/49.6} & \textbf{42.4/18.0} & 2.4/5.7/19.2 \\
        + rm. prev. resp. & - & - & \textbf{2.9/8.8}/21.7 \\
        \bottomrule
    \end{tabular}
    \captionof{table}{Zero-shot performance on highlight detection, temporal video grounding, and dense video captioning. All models uses 7B LLMs.} \label{tab:grounding_and_captioning}

\end{minipage}

\end{figure*}

The data distribution of \datasetname is shown in \cref{fig:data_distribution}. Note that this dataset only contains 109k examples, which is relatively small compared to modern post-training datasets like \cite{Li2023MVBenchAC,Li2024LLaVAOneVisionEV,Wang2024HawkEyeTV}. 
The reason is that due to computational resource constraints, we plan to demonstrate the feasibility of our proposed video-text duet interaction format by fine-tuning a state-of-the-art VideoLLM. We assume that the used backbone model already possesses enough video comprehension capabilities. By using a small dataset, we aim to train this model to efficiently adopt this new interaction with minimum catastrophic forgetting of its existing abilities.


\section{Experiments} \label{sec:experiments}

\paragraph{Implementations}
\modelname is initialized with LLaVA-OneVision \cite{Li2024LLaVAOneVisionEV}. We train the model on \datasetname for one epoch. The training takes about one day on a node with 8 Tesla V100 GPUs, and the inference runs on 1 Tesla V100 GPU. More implementation details are listed in \cref{sec:hyperparameters}.

\paragraph{Baselines}
As \modelname mainly focuses on time-sensitive video tasks, we use the following baselines that are able to represent time spans in videos by different representation formats: TimeChat \cite{Ren2023TimeChatAT}, VTimeLLM (7B) \cite{Huang2023VTimeLLMEL}, HawkEye \cite{Wang2024HawkEyeTV}, VTG-LLM \cite{Guo2024VTGLLMIT}, and VideoLLM-Online \cite{Chen2024VideoLLMonlineOV}. For VideoLLM-Online, we experimented with $\theta \in \{0.5, 0.6, 0.7, 0.8\}$ as suggested in their paper and report the best results (0.8 for both dense video captioning and MAGQA).

Since the initialization of \modelname is stronger than that of the baselines, for a fair comparison we also conduct a controlled experiment in which the only difference is the interaction format. Specifically, we use the same initialization model (LLaVA-OneVision), training data (\datasetname) and training schedule, but reformat the data to the respective interaction formats and video segment representation formats used by TimeChat and VTimeLLM to train two baseline models. We name these models as LLaVA-OV-TC and LLaVA-OV-VT.

\subsection{Highlight Detection and Temporal Video Grounding}

We use highlight detection and temporal video grounding to evaluate the performance of the relevance head of \modelname.
Baseline models are required to generate a list of float numbers to represent the relevance score for each clip in QVHighlights \cite{Lei2021QVHighlightsDM}, and a start and end time for the relevant video span in Charades-STA. However, for LLaVA-OV-TC and LLaVA-OV-VT, despite using different prompts as input, we were still unable to instruct the model to output a sequence of scores as in \cite{Ren2023TimeChatAT}. Therefore, we follow the method of Charades-STA to instruct the model to output a related span, and assign the score to 1 for clips within this span and 0 otherwise. \modelname uses the relevance score min-max normalized to $[0, 1]$ as the score in QVHighlights, and to classify whether this frame is relevant and calculate frame-level IoU in Charades-STA.

Since the relevance head provides a relevance score immediately after each frame, its prediction cannot leverage the context from subsequent video frames. To mitigate this limitation, we smooth the relevance score sequence. Specifically, we set each frame's smoothed relevance score as the mean value of its original score, the relevance scores of the preceding $w$ frames and the following $w$ frames, where $w$ is the window size. We set $w = 2$ for QVHighlights and $w = 6$ for Charades-STA. Results are shown in \cref{tab:grounding_and_captioning}. We observe that, compared to the baselines, \modelname exhibits a significantly greater improvement in performance on QVHighlights. This indicates that traditional VideoLLMs struggle with generating a long sequence of relevance scores using a text-based form or identifying multiple related video segments in its text-based responses, whereas \modelname's approach of directly assigning relevance scores to each frame circumvents this issue. For VideoLLM-Online, we instruct it to reply with ``start'' / ``end'' at the start / end time of the target clip following the examples given in its paper but it does not follow the instructions despite trying different wordings, so we are not able to report its performance.

\paragraph{$w$ is robust to different values}
Though the $w$ is empirically set for the results in \cref{tab:grounding_and_captioning}, we also find that within a fairly large range of $w$, the performance of \modelname is robust and consistently outperforms all baseline models. Detailed results are listed in \cref{sec:hyperparameter_sensitivity}.

\subsection{Dense Video Captioning} \label{exp:sec_dvc}

We test dense video captioning performance on YouCook2 \cite{Zhou2017TowardsAL}, a challenging task that requires models to output the caption, start point and end point for about 8 steps in a minutes-long cooking video. Baseline models output the start time, end time and caption for each step in the text-based form. 
For \modelname, since this task requires the model to continuously identify important actions from the video and output periodically, we employ a heuristic method to determine whether a model response should be output after each frame (\texttt{need\_response} function in \cref{sec:inference}). We sum up the informative score for each frame as the video plays. When the sum reaches a threshold $s$ (we set $s=2$), the model generates a response right after this frame as the caption for that step, and then we reset the sum to $0$ to start a new round of sum.

However, \modelname cannot directly predict when a step starts or ends just by this video-text duet interaction format, as the model is unable to determine whether a frame is the beginning of a step without observing enough subsequent content. To get the start and end time for each step as required by this task, we adopt a simple workaround: we use the time of the previous response and the current response as the start time and end time for a step. If two adjacent steps have the same caption, we merge them into one step. This workaround is also applied on VideoLLM-Online.

It has been a long-lasting problem that LLMs tend to repeat previously-generated content \cite{Xu2022LearningTB}, and we find that this problem is especially severe in dense video captioning. It indicates that VideoLLMs are probably generating captions relying on text shortcuts rather than the video content. We have attempted common solutions such as repetition penalty \cite{Keskar2019CTRLAC}, which though is still sub-optimal.
Since the responses from \modelname are separated across multiple turns, we find that simply removing previously generated turns from the context (``rm. prev. resp.'' for short) by not appending their attention keys and values to the KV Cache alleviates this issue, leading to a significant improvement in performance. However, this simple trick is not applicable to ``whole-video'' format baselines, as if the latest words are removed from the KV Cache, it will remain the same as before generating the latest words and the model will generate the same words again, despite some minor changes due to random sampling. In contrast, for \modelname new video contents continuously bring new KV Cache and drive the conversation forward.

As shown in \cref{tab:grounding_and_captioning}, \modelname does not show significant improvements on F1 metric, likely due to the simple solution we use to derive the start and end time based on responses. Even so, the CIDEr and CODA\_c metric (inaccurate predicted time spans can have negative effects on these metrics) of \modelname is still higher than all baselines, indicating that \modelname outperforms baselines in terms of text quality, possibly due to its facilitation to information retrieval discussed in \cref{sec:intro}.

\paragraph{$s$ is robust to different values}
We also find that the threshold $s$ is quite robust across a wide range of from 1 to 3, and we can use different $s$ to suit various downstream tasks especially in such zero-shot setting. Detailed results are listed in \cref{sec:hyperparameter_sensitivity}.

\begin{table*}[h!]
    \centering
    \small
    \setlength\tabcolsep{3pt}
    \begin{tabular}{c|c|c|c|c|c|c|c}
        \toprule
        \multirow{2}{*}{\makecell[c]{\\ Model}} & \multirow{2}{*}{\makecell[c]{\\ Real- \\ Time?}} & \multicolumn{3}{c}{original test set} & \multicolumn{3}{|c}{5-time prolonged video test set} \\
        \cline{3-8}
        & & \makecell[c]{In-Span Score \\ LLaMA/GPT} & \makecell[c]{\# turns (w/o. \\/ w/. dedup)} & \makecell[c]{time per \\ example} & \makecell[c]{In-Span Score \\ LLaMA/GPT} & \makecell[c]{\# turns (w/o. \\/ w/. dedup)} & \makecell[c]{time per \\ example} \\ 
         \hline
        \multicolumn{4}{l}{\textit{Baselines}} \\
        \hline
        LLaVA-OV-TC & \usym{2718} & 2.77/2.64 & 4.1/2.2 & \textbf{1.00} & 1.67/1.62 & 7.6/2.4 & 1.00 \\
        LLaVA-OV-VT & \usym{2718} & 2.54/2.42 & 4.1/3.1 & 1.06 & 1.64/1.60 & 10.2/3.4 & \textbf{0.99} \\
        VideoLLM-Online & \usym{2714} & 1.33/1.26 & 1.3/1.1 & \textcolor{lightgray}{0.44}$^\ast$ & - & - & - \\ 
        \hline
        \multicolumn{4}{l}{\textit{\modelname(Ours)}} \\
        \hline
        $t=0.6$ & \usym{2714} & 2.46/2.33 & 13.7/4.0 & 1.90 & 1.83/1.73 & 22.3/7.0 & 1.04 \\
        $t=0.5$ & \usym{2714} & 2.77/2.61 & 18.4/5.3 & 2.36 & 2.16/2.02 & 31.2/9.8 & 1.45 \\
        $t=0.4$ & \usym{2714} & 3.00/2.81 & 23.0/6.6 & 2.75 & 2.44/2.28 & 41.7/13.0 & 2.17 \\
        $t=0.3$ & \usym{2714} & \textbf{3.13/2.93} & 27.0/7.6 & 2.90 & \textbf{2.63/2.45} & 52.8/16.5 & 2.62 \\
        \bottomrule
    \end{tabular}
    \caption{Results on the test set of Shot2Story-MAGQA-39k with the rm. ass. turns method used. For the ``time per example'' column, the time used by ``LLaVA-OV-VT'' is set to 1, and the times for other rows are set as multiples of the time used by ``LLaVA-OV-TC''. $^\ast$: Inference time of VideoLLM-Online is changed to gray and de-emphasized as it only generates one reply immediately after the question and is hardly helpful for answering the question, and thus we no longer evaluate it on the 5-times prolonged video test set.}
    \label{tab:magqa_main_results}
\end{table*}

\subsection{Multi-Answer Grounded Video QA} \label{sec:exp_magqa}
To align closely with the widely-used streaming video comprehension scenario, we propose MAGQA that requires a model to generate answers at multiple necessary positions of a video.
Different from conventional Video QA in which one question corresponds to only one answer, In MAGQA, a question corresponds to multiple turns of answers, and these turns are derived from different video segments. Therefore, this task requires the response to be accurate and in time.
Though under the video-text duet interaction format users may raise arbitary number of questions at any time, to ensure the feasibility of evaluation, in this experiment we assume that the user raises only one question at the beginning of the video, and leave the extension to multiple questions as future work.

As this task is a newly-proposed one, we introduce an ``in-span score'' metric, which uses LLMs to calculate the average similarity of pred answers and gold answers that falls into the same time span of response, to evaluate both the correctness and timeliness of model responses. A detailed description of this metric is in \cref{sec:detail_in_span_score}. To prevent reproducibility issues due to potential changes of OpenAI API, besides GPT-4o-2024-08-06 \cite{openai2024gpt4o}, we also report the in-span score obtained using LLaMA 3.1 70B Instruct \cite{Dubey2024Llama3} to calculate pred-gold similarities.


As MAGQA requires the answers to be both informative and related to the question, we set \texttt{need\_response} as: if the sum of informative score and relevance score of a frame is larger than a threshold $t$, then the model needs to generate a response right after this frame. We also use the ``rm. prev. resp.'' method in dense video captioning task introduced in \cref{exp:sec_dvc}.
As baseline models are not capable of generating responses at specific positions in the video, we employ an output format the same as dense video captioning, \textit{i.e.,} output the start time, end time, and predicted text for each turn after watching the entire video in both training and testing, and use the average of the start and end time as the response time. We also observe that for some cases the baseline models directly give one answer instead of generating multiple replies and their corresponding time spans, and we do not count these examples into the metrics when reporting results.
\textbf{Note that this is a significantly simplified requirement than that of \modelname}, as the MAGQA task simulates streaming video comprehension application scenario, which requires the model to respond as soon as the video plays to segments relevant to the question, which ensures that users can see the responses timely, rather than waiting until the entire video concludes before generating replies.

\paragraph{\modelname has better performance than baselines and provides real-time replies.}
Results on the test set of Shot2story-MAGQA-39k are shown in the lelf half of \cref{tab:magqa_main_results}. We provide results for different $t$ as it represents a trade-off between inference time and performance: as $t$ decreases from $0.6$ to $0.3$, the performance of \modelname's real-time replies continuously rises and outperforms baselines with a simplified setting of providing non-real-time replies after watching the entire video. However, this is achieved at a cost of generating lots of duplicate replies with more inference time.

\paragraph{\modelname performs much better than baselines on longer videos.} 
Since the average video length of the test set of Shot2story-MAGQA-39k is only 16.9 seconds, to demonstrate \modelname’s real-time QA capabilities on longer videos we use a simple approach to make videos in the test set longer: we splice the video with 4 other videos randomly selected from the test set in random order to prolong the video to approximately 5 times longer by padding with videos irrelevant to the question. Results on the prolonged videos are shown in the right half of \cref{tab:magqa_main_results}. When the videos are long, it becomes harder for baseline models to output correct time spans for the answers which results in low in-span scores, while \modelname is more likely to generate correct answers at the right time.

\begin{table}
    \begin{minipage}{0.46\columnwidth}
    \centering
    \small
    \begin{tabular}{c|c}
    \toprule
    Model & Acc \\
    \hline
    Flash-VStream & 1.96 \\
    VLLM-Online & 3.92 \\
    Dispider & 25.34 \\
    \modelname & \textbf{29.44} \\
    \bottomrule
    \end{tabular}
    \caption{Performance on the Proactive Output task of StreamingBench.} \label{tab:streamingbench}
    \end{minipage}%
\hfill 
    \begin{minipage}{0.5\columnwidth}
    \centering
    \small
    \begin{tabular}{c|c}
    \toprule
    Model & YouCook2 \\
    \hline
    \modelname & \textbf{2.9/8.8/21.7} \\
    \hline
    \makecell[c]{w/o rand. \\ resp. pos.} & 2.1/7.3/19.0 \\
    \hline
    \makecell[c]{w/o multi \\ informative} & \textbf{2.9}/8.0/16.5 \\
    \bottomrule
    \end{tabular}
    \caption{Ablation study on training methods.} \label{tab:train_ablation}
    \end{minipage}
\end{table}

\subsection{Proactive Output on StreamingBench}
To further demonstrate the timeliness of the replies of \modelname, we also report results on the Proactive Output task of StreamingBench \cite{Lin2024StreamingBenchAT}. StreamingBench evaluates VideoLLMs in real-time, streaming video understanding tasks. Specifically, for the ``Proactive Output'' task, a question is considered as correctly answered if a reply is raised by the model within two seconds when a certain scene that contains the answer appears. Results in \cref{tab:streamingbench} show that \modelname outperforms all Streaming or Proactive MLLMs \cite{Zhang2024FlashVStreamMR, Chen2024VideoLLMonlineOV, Qian2025DispiderEV}. Refer to \cref{sec:streamingbench_more_results} for more details and baselines.

\subsection{Ablation Studies}
We conduct ablation studies on YouCook2 dense video captioning to assess two empirical yet important findings for effectively training the informative head in data construction: randomly inserting the response at a position from 50\% to 75\% of the corresponding video segment (rand. resp. pos.), and setting informative head’s label to \textit{TRUE} for all frames between 50\% of the segment and the response time (multi informative). When ``rand. resp. pos.'' is disabled, the response is always inserted at the end of the corresponding segment. When ``multi informative'' is disabled, only the informative label of the frame right before the response is set as \textit{TRUE}. As illustrated in \cref{tab:train_ablation}, disabling either method negatively impact \modelname's performance, which shows the importance of carefully handling the response time and informative labels.

\section{Conclusion}
In this paper, we first formalize the video-text duet interaction format. We collect \datasetname for training models to follow the video-text duet interaction format. Based on \datasetname we train \modelname, a model with significant improvements on various time-sensitive tasks and is able to automatically decide when to response in a real-time manner. We believe such improvements can be a substantial step towards building powerful and useful video comprehension applications.

\section*{Limitations}
We acknowledge that there is much room for improvement which should be addressed in future research:
(1) Some hyperparameters (\textit{e.g.,} the \texttt{need\_response} criterion) are required during inference. However, we have shown that this criterion is quite robust across different thresholds. 
(2) Information from subsequent frames is not incorporated when generating in-time responses for the current frame, especially for the live-streaming video that indeed has unpredictable future frames. It can be crucial in some scenarios, such as determining the start of an action.
(3) Slow inference speed. A better inference process is needed for avoid generating duplicate responses.
(4) Real-time response datasets with longer live-streaming videos are required to be collected to better fit the real-world application scenarios.

\section*{Acknowledgement}
This work is supported in part by the State Key Laboratory of General Artificial Intelligence.



\bibliography{custom}

\appendix
\section{Data Quality Check of Shot2Story-MAGQA-39k} \label{sec:data_quality}
We sample 100 examples (with 290 answers) from our test set for manual quality assessment. Among the sampled examples, we find 1 example with a question unanswerable from the video, 5 examples have 6 answers (2.1\%) that contradict the video content, and 5 examples have 7 answers (2.4\%) unrelated to the question. Overall, manual quality assessment shows that above 95\% data of our test set belongs to the high quality, which confirms the potential value of using Shot2Story-MAGQA-39k to benchmark models. The reason for the high quality is when the video captions are provided, generating questions and answers based on these text captions is a very simple task for advanced LLMs like GPT4o.
However, we also find that in 21 examples, the video contains additional information that is not covered in the answers. This is because some questions are very general, like "What scene is the video displaying?", and describing scenes in videos elaborately has been a long-lasting challenge for annotating video datasets.

\section{Details of Training and Inference}
\subsection{Data Reformat Process of \datasetname}
\label{sec:detail_reformat_process}
In \cref{sec:data_dense_captioning} we briefly introduced how the annotations for offline dense captioning / QA are converted into image-text interleave interactive format for training \modelname. Here we elaborate more details and the reasons of this design:

\paragraph{Choices of insertion}
We randomly sample a position from 50\% to 75\% time duration for the corresponding video segment, and insert the caption at that position as a model response. Here we introduce some randomness in the insertion position to prevent the model from developing a bias or a shortcut such as responses can only be generated at some specific positions.
The earliest and latest time for inserting responses, \textit{i.e.,} at the 50\% and 75\% place of segment duration, are empirically chosen, as it works well in our preliminary study. We avoid inserting responses too early like in the first half of duration, because it is unfeasible to generate responses related to this video segment at a very starting point. It is reasonable that some further observations are required to gain a more comprehensive understanding of it. We also avoid inserting responses too late like in the last one-fourth duration, as we hope the model to output a response as soon as it has a sufficient understanding of the segment, rather than wait until the disappearance of the segment. It thereby improves the timeliness of the whole interaction between users and videos, especially when the user can still watch the segment as well as perceive the content of the model response talking about it.

\paragraph{Creating informative labels}
We also create labels for the informative head in dense captioning tasks. According to the previous paragraph, the model can not have a comprehensive understanding of this video segment until it has viewed a sufficient portion of the segment (50\% in this case). Meanwhile, once the caption has been generated as model response, we assume that the remaining frames in this video segment no longer provide new information that is not covered in the caption. Therefore, we set the informative head’s label to \textit{TRUE} for frames between 50\% of this segment and the insertion point of the response, and set labels to \textit{FALSE} for the other frames.

\subsection{Training Hyperparameters} \label{sec:hyperparameters}
LLaVA-OneVision uses SigLIP-Large \cite{Zhai2023SigmoidLF} as the vision encoder, and converts an image with $384 \times 384$ into $24 \times 24 = 576$ tokens. 
In the official settings of LLaVA-OneVision \cite{Li2024LLaVAOneVisionEV}, when encoding videos, the visual tokens corresponding to each frame are spatially downsampled to $12 \times 12 = 144$ tokens using a pooling operation with a size of 2. However, this number of tokens is also too large when training and inference with long videos. To address this, we further modified the pooling size to 4, resulting in $7 \times 7 = 49$ tokens per frame.

We set the maximum number of frames sampled from each video to 120 in the training process, which is constrained by the memory of our GPUs. The sampling frame rates are set to different numbers for different video sources to ensure that for the vast majority ($\textgreater$90\%) of videos, video length (in seconds)  $\div$ sampled frame per second (fps) $\le 120$. For the videos that are too long, we only keep the first 120 frames (and the conversation turns that are inserted within the first 120 frames), and discard the subsequent contents. Specifically, the sampled frame per second (fps) is set as: $2$ for videos from Shot2Story \cite{Han2023Shot2Story20KAN} and DiDeMo \cite{Hendricks2017LocalizingMI}, $0.5$ for COIN \cite{Tang2019COIN} and QueryD \cite{Oncescu2021QUERYDAV}, and 0.33 for $\text{HiREST}_{grounding}$ \cite{Zala2023HierarchicalVR}.

The projector, the relevance head, the informative head and LoRA \cite{hu2022lora} weights of the LLM (add to all attention proj. layers and FFN layers) are trained, while other parameters of the model are frozen. More training hyperparameters are listed in \cref{tab:hyperparameters}.

\subsection{Pseudo Code of the Inference Process} \label{sec:pesudo_code_inference}
We provide a python-style pseudo code of the inference process in \cref{alg:inference}.

\begin{figure}[t]
\lstset{ 
    basicstyle=\footnotesize\ttfamily,
    keywordstyle=\bfseries,
    texcl=true,
    frame=single,
    captionpos=b
}
\noindent
\begin{minipage}{\columnwidth}
\begin{lstlisting}[caption={Inference Process of \modelname}, label={alg:inference}]
# Input: 
#  system_prompt
#  video: list of frames
#  fps: frames per second to sample from video
#  user_turns: list of (time, text) sorted by time
# Output:
#  model_turns: generated list of (time, text)

model_turns = []
v_inf_list, v_rel_list = [], []
kv_cache = model(system_prompt)
time = 0
for frame in video:
  if len(user_turns) and time>=user_turns[0].time:
    kv_cache = model(kv_cache, user_turns[0].text)
    user_turns = user_turns[1:]
  kv_cache, v_inf, v_rel = model(kv_cache, frame)
  v_inf_list.append(v_inf)  # informative score
  v_rel_list.append(v_rel)  # relevance score
  if need_response(v_inf_list, v_rel_list):
    kv_cache, response = model.generate(kv_cache)
    model_turns.append((time, response))
  time += 1 / fps
\end{lstlisting}
\end{minipage}
\end{figure}

\subsection{Inference Settings}
Videos from different sources are also sampled with different fps during inference. Specifically, we set the maximum number of frames sampled from each video to 400, and fps to 2 for videos from Shot2Story \cite{Han2023Shot2Story20KAN} and Charades-STA \cite{Gao2017TALLTA}, 1 for videos from QVHighlights \cite{Lei2021QVHighlightsDM}, and 0.5 for videos from YouCook2 \cite{Zhou2017TowardsAL}. For a few videos in YouCook2 that are even longer than $400 \text{(frames)} \div 0.5 \text{(fps)} = 800$ seconds, we uniformly sample $400$ frames from this video to ensure that information from the latter part of the video is not truncated. This inference setting is consistent across \modelname, LLaVA-OV-TC, and LLaVA-OV-VT.

\begin{table}[]
    \centering
    \begin{tabular}{c|c}
        \toprule
        Hyper-parameter & value \\
        \hline
        \texttt{batch\_size} & 1 \\
        \texttt{gradient\_acc\_steps} & 8 \\
        \texttt{learning\_rate} & 2e-5 \\
        \texttt{warmup\_ratio} & 0.05 \\   
        \texttt{lora\_r} & 16 \\
        \texttt{lora\_alpha} & 32 \\
        \texttt{attn\_implementation} & sdpa \\
        \bottomrule
    \end{tabular}
    \caption{Hyper-parameters used for training \modelname.}
    \label{tab:hyperparameters}
\end{table}


\subsection{Details of the In-Span Score} \label{sec:detail_in_span_score}
Suppose the model prediction has $P$ answers, each answer has a prediction time $time_p$ and prediction text $pred_p$, $p=1,2,\dots,P$. The ground truth has $Q$ answers, each answer has a ground truth start time $start_q$, a ground truth end time $end_q$, and a ground truth text $gold_q$, $q=1,2,\dots,Q$. First, we use an LLM to calculate a relevance score from 1 to 5 between each answer in prediction $pred_p$ and ground truth $gold_q$: $S = \{s_{p,q}\} \in \mathcal{R}^{P \times Q}$.
For each ground truth answer $q$, we select the predicted answers with predicted time in ground truth time span: $\mathcal{P}_q = \{ p \mid time_p \in [start_q, end_q] \}$, and use the average score between the ground truth answer and the selected predicted answers as the score for this ground truth answer: $score_q = \frac{1}{|\mathcal{P}_q|} \sum_{p \in \mathcal{P}_q} s_{p,q} \quad \text{if } |\mathcal{P}_q| > 0$. If $|\mathcal{P}_q| = 0$  (no predicted answer falls in this ground truth span), $score_q$ is set to $1$.  Finally, we calculate the average score of all ground truth answers as the final in-span score of this example: $in\_span\_score = \frac{1}{|Q|} \sum_{q=1}^{|Q|} score_q$.

\section{More Experimental Results}
\subsection{Hyperparameter Sensitivity} \label{sec:hyperparameter_sensitivity}
\begin{figure}
    \centering
    \includegraphics[width=0.45\linewidth]{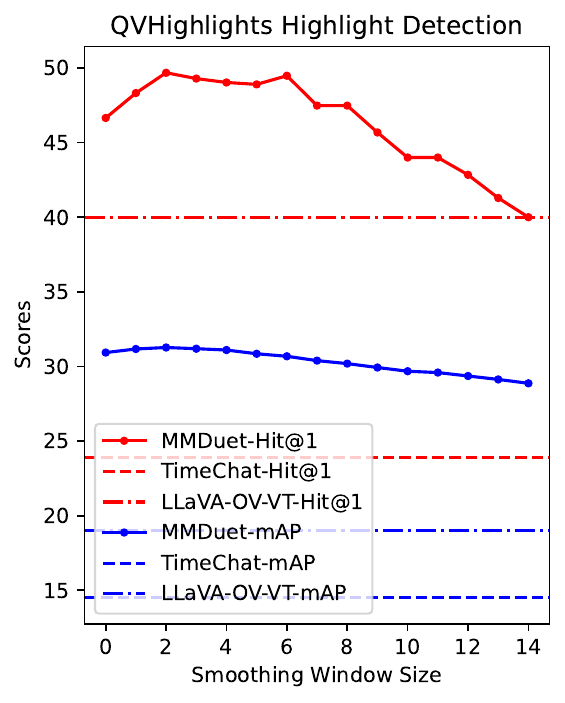}
    \includegraphics[width=0.45\linewidth]{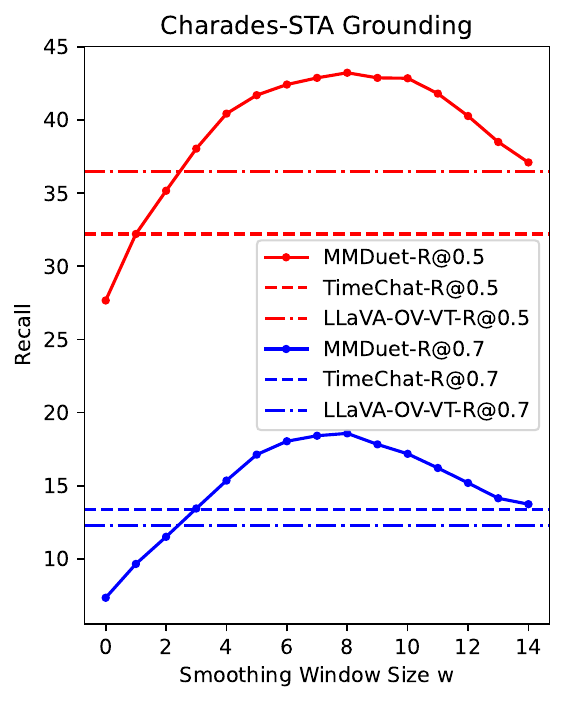}
    \caption{Performance on temporal video grounding and highlight detection with different $w$.}
    \label{fig:smoothing}
\end{figure}

\begin{figure}
    \centering
    \includegraphics[width=\linewidth]{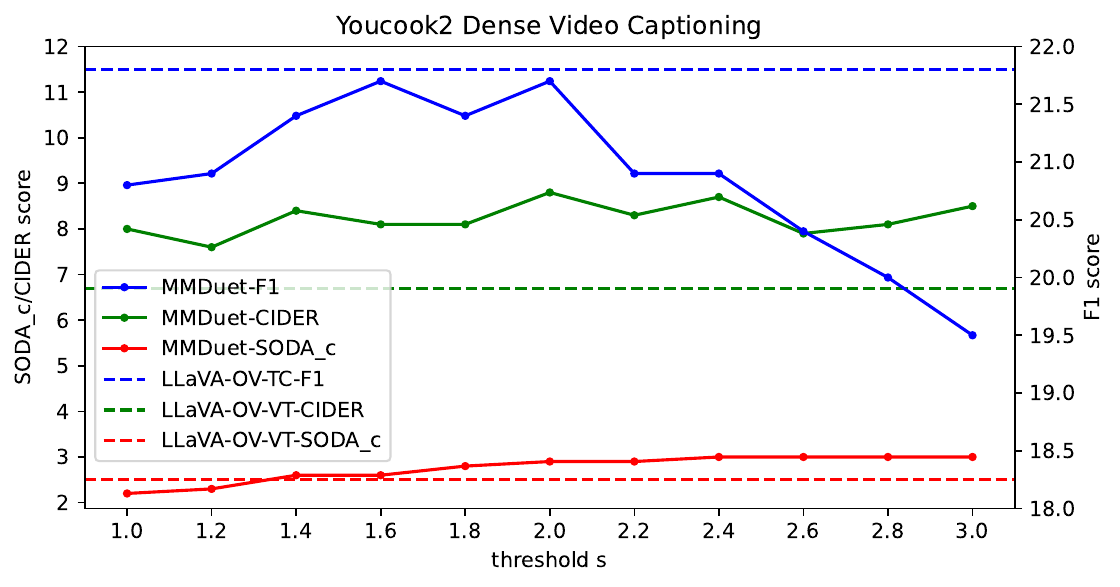}
    \caption{Performance on dense video captioning with different $s$.} \label{fig:captioning_s}
\end{figure}

We list the experiments using different window size $w$ for temporal grounding in \cref{fig:smoothing} and threshold $s$ for dense captioning in \cref{fig:captioning_s}.

\subsection{Details of the Proactive Output Experiment} \label{sec:streamingbench_more_results}
More results and baselines are listed in \cref{tab:streamingbench_more_results}.
For results of models without streaming abilities (Proprietary MLLMs \& Open-Sourced VideoLLMs), we follow the evaluation method of \cite{Lin2024StreamingBenchAT} and \cite{Qian2025DispiderEV}: We gradually extend the input video one second at a time and ask the model with the question “Is it the right time to output?”. If the model responds with “Yes.”, this moment is recorded as the predicted output timestamp.
For \modelname, we use the time of the first reply after the user question is input as the predicted output timestamp. For examples that \modelname does not provide any reply at all, we consider them as failing cases and the difference between ground truth output time and predicted output time is recorded as $+\infty$.

\begin{table}[]
    \centering
    \small
    \setlength{\tabcolsep}{3pt} 
    \begin{tabular}{c|c||c|c}
    \toprule
    Model & Acc & Model & Acc \\
    \hline
    \multicolumn{4}{l}{\textit{Proprietary MLLMs}} \\
    \hline
    Gemini 1.5 pro & 45.10 & GPT-4o & 56.86 \\
    Claude 3.5 Sonnet & 64.71 & & \\
    
    \hline
    \multicolumn{4}{l}{\textit{Open-Sourced VideoLLMs}} \\
    \hline
    LLaVA-OV & 29.55 & Qwen2-VL & 22.73 \\
    MiniCPM-V 2.6 & 22.22 & LLaVA-NeXT-Video & 18.18 \\
    InternVL2 & 40.91 & LongVA & 15.91 \\
    
    \hline
    \multicolumn{4}{l}{\textit{Streaming MLLMs}} \\
    \hline
    Flash-VStream & 1.96 & VideoLLM-Online & 3.92 \\
    Dispider & 25.34 & & \\
    \hline
    \modelname $t=0.3$ & 29.44 & \modelname $t=0.4$ & 31.85 \\
    \modelname $t=0.5$ & 26.61 & \modelname $t=0.6$ & 18.95 \\
    \bottomrule
    \end{tabular}
    \caption{Performance of more baselines and \modelname on the Proactive Output task of StreamingBench with different $t$.}
    \label{tab:streamingbench_more_results}
\end{table}

\section{Example Inputs for Each Task in \datasetname} \label{sec:example_input}
Example inputs for each task for training and inference are listed in \cref{tab:input_exapmles}. The dense video captioning user input is selected from one of the following sentences:

\noindent\rule{\linewidth}{0.5pt} 
{
\small
\noindent Please concisely narrate the video in real time.

\noindent Help me to illustrate my view in short.

\noindent Please simply describe what do you see.

\noindent Continuously answer what you observed with simple text.

\noindent Do concise real-time narration.

\noindent Hey assistant, do you know the current video content? Reply me concisely.

\noindent Simply interpret the scene for me.

\noindent What can you tell me about? Be concise.

\noindent Use simple text to explain what is shown in front of me.

\noindent What is the action now? Please response in short.
}
\newline\noindent\rule{\linewidth}{0.5pt} 

The temporal video grounding user input is selected from one of the following sentences (where ``\%s'' denotes the caption to localize):

\noindent\rule{\linewidth}{0.5pt} 
{
\small
\noindent \%s
\noindent What segment of the video addresses the topic '\%s'?

\noindent At what timestamp can I find information about '\%s' in the video?

\noindent Can you highlight the section of the video that pertains to '\%s'?

\noindent Which moments in the video discuss '\%s' in detail?

\noindent Identify the parts that mention '\%s'.

\noindent Where in the video is '\%s' demonstrated or explained?

\noindent What parts are relevant to the concept of '\%s'?

\noindent Which clips in the video relate to the query '\%s'?

\noindent Can you point out the video segments that cover '\%s'?

\noindent What are the key timestamps in the video for the topic '\%s'?
}
\newline\noindent\rule{\linewidth}{0.5pt} 

\begin{table*}
\centering
\begin{tabular}{l|l}
    \toprule
    \makecell[l]{Dense\\Video\\Captioning} & \makecell[{{p{12cm}}}]{\texttt{<im\_start>}system\\
    A multimodal AI assistant is helping users with some activities. Below is their conversation, interleaved with the list of video frames received by the assistant. \texttt{<im\_end>}\\
    \texttt{<im\_start>}user \\
    (A Dense Video Captioning Query)\texttt{<im\_end>}\\
    \texttt{<im\_start>}stream \\
    \texttt{<frame><frame><frame> \ldots <im\_end>} \\
    \texttt{<im\_start>}assistant \\
    A person pulls a knife from a black bag.\texttt{<im\_end>} \\
    \texttt{<im\_start>}stream \\
    \texttt{<frame><frame><frame> \ldots <im\_end>} \\
    \texttt{<im\_start>}assistant \\
    A man in a hat and red clothes speaks with a dagger, and a tree behind him.\texttt{<im\_end>} \\
    \texttt{<im\_start>}stream \\
    \texttt{<frame><frame><frame> \ldots <im\_end>} \\
    \ldots (More stream and assistant turns)
    } \\
    \hline
    MAGQA & \makecell[{{p{12cm}}}]{\texttt{<im\_start>}system\\
    A multimodal AI assistant is helping users with some activities. Below is their conversation, interleaved with the list of video frames received by the assistant. \texttt{<im\_end>}\\
    \texttt{<im\_start>}stream\\
    \texttt{<frame><frame><frame> \ldots <im\_end>} \\
    \texttt{<im\_start>}user \\
    What happens during the basketball game?\texttt{<im\_end>}\\
    \texttt{<im\_start>}stream \\
    \texttt{<frame><frame><frame> \ldots <im\_end>} \\
    \texttt{<im\_start>}assistant \\
    Several players in white jerseys are celebrating by high-fiving each other.\texttt{<im\_end>} \\
    \texttt{<im\_start>}stream \\
    \texttt{<frame><frame><frame> \ldots <im\_end>} \\
    \texttt{<im\_start>}assistant \\
    A player in a white jersey makes a successful shot.\texttt{<im\_end>} \\
    \texttt{<im\_start>}stream \\
    \texttt{<frame><frame><frame> \ldots <im\_end>} \\
    \ldots (More stream and assistant turns)
    } \\
    \hline
    \makecell[l]{Temporal\\Video\\Grounding} & \makecell[{{p{12cm}}}]{\texttt{<im\_start>}system\\
    A multimodal AI assistant is helping users with some activities. Below is their conversation, interleaved with the list of video frames received by the assistant. \texttt{<im\_end>}\\
    \texttt{<im\_start>}user \\
    (A Temporal Video Grounding Query)\texttt{<im\_end>}\\
    \texttt{<im\_start>}stream \\
    \texttt{<frame><frame><frame> \ldots <im\_end>} \\
    } \\
    \bottomrule
\end{tabular}
\caption{Input examples of different tasks during the training and evaluation phase of \modelname.} 
\label{tab:input_exapmles}
\end{table*}

\section{Qualitative Study}
\begin{figure*}
    \centering
    \includegraphics[width=\textwidth]{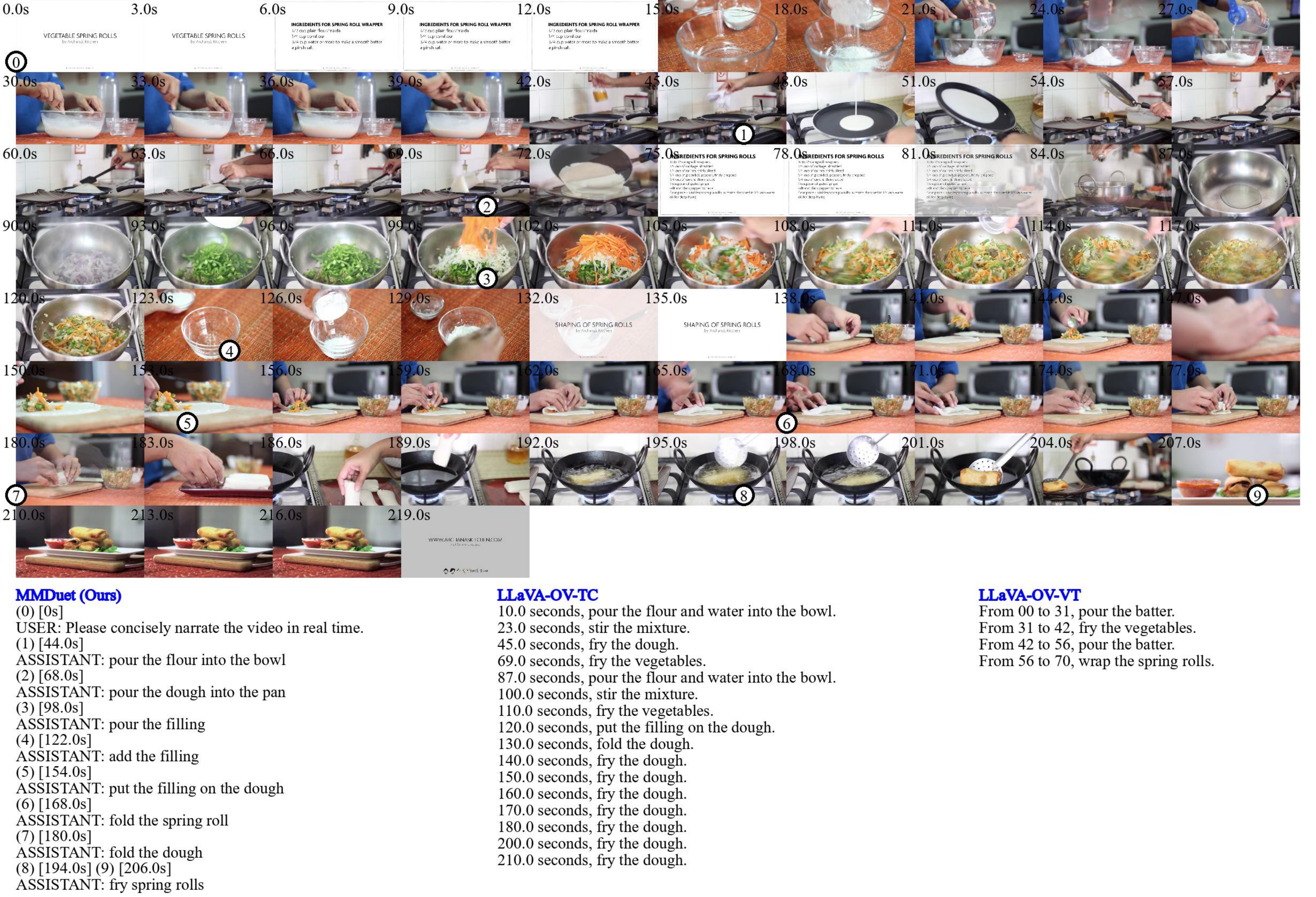}
    \caption{An example of dense video captioning with \modelname, LLaVA-OV-TC and LLaVA-OV-VT.}
    \label{fig:dvc_example1}
\end{figure*}

\begin{figure*}
    \centering
    \includegraphics[width=\textwidth]{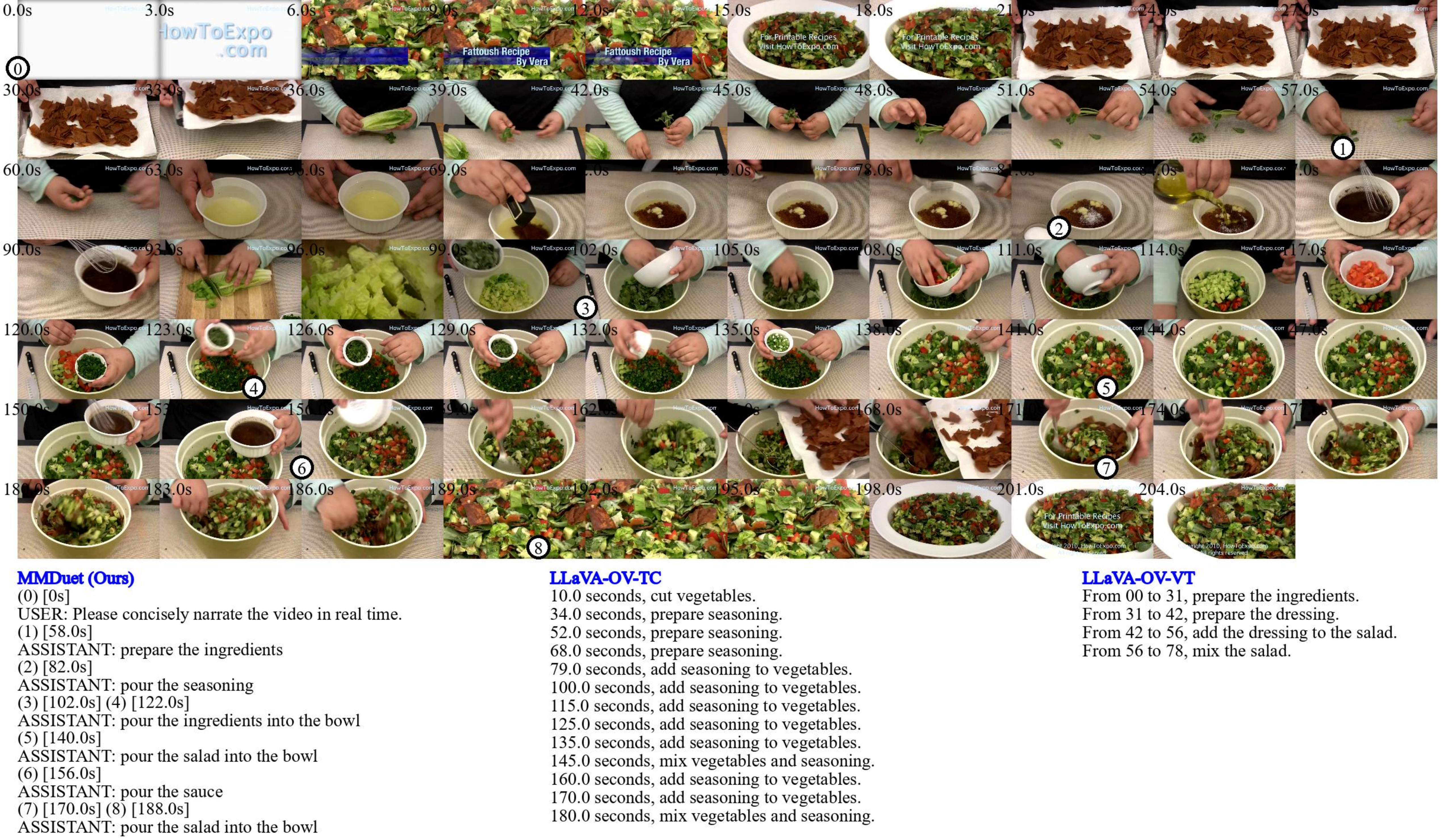}
    \caption{An example of dense video captioning with \modelname, LLaVA-OV-TC and LLaVA-OV-VT.}
    \label{fig:dvc_example2}
\end{figure*}

\begin{figure*}
    \centering
    \includegraphics[width=\textwidth]{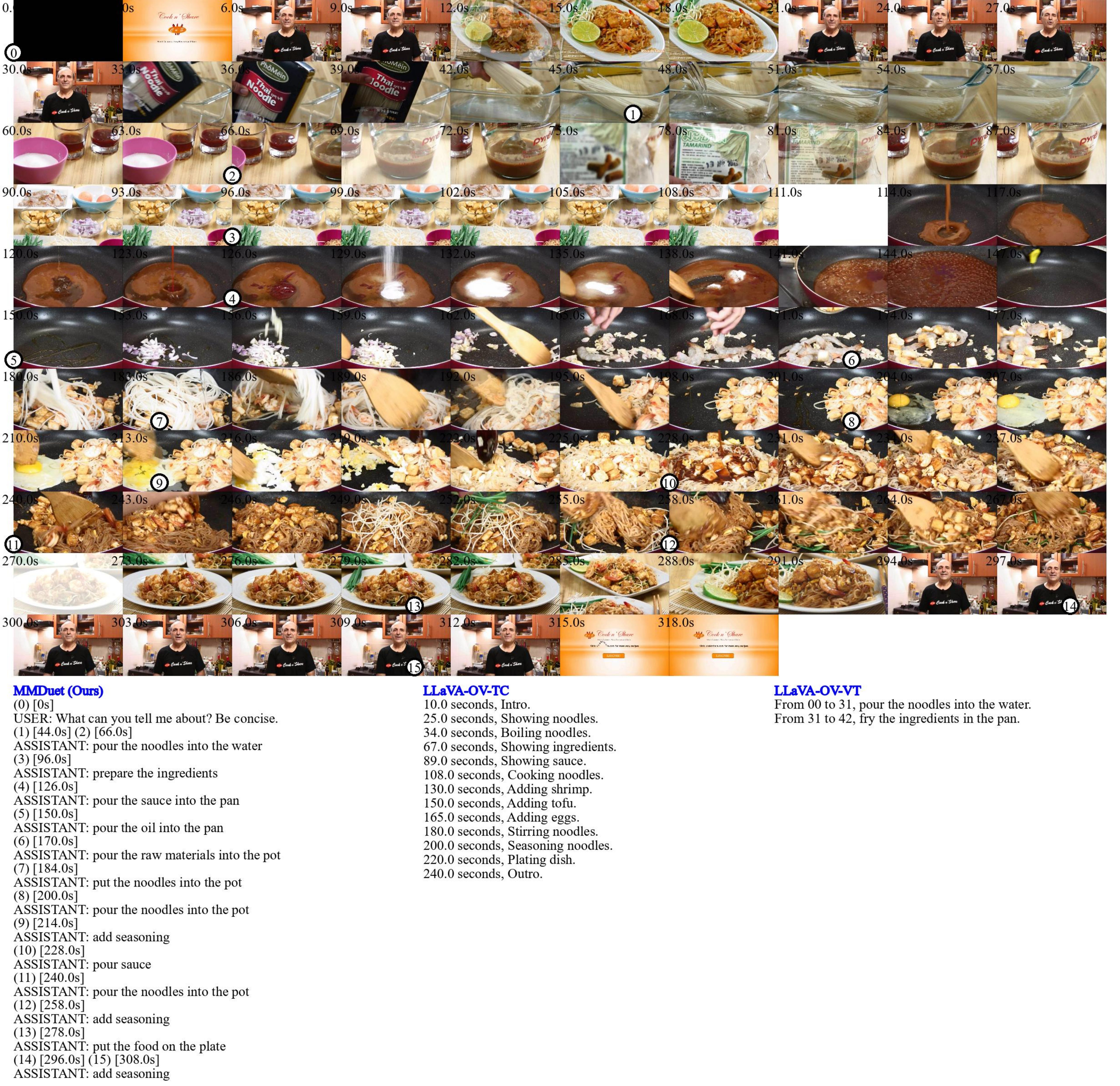}
    \caption{An example of dense video captioning with \modelname, LLaVA-OV-TC and LLaVA-OV-VT.}
    \label{fig:dvc_example3}
\end{figure*}

\begin{figure*}
    \centering
    \includegraphics[width=\textwidth]{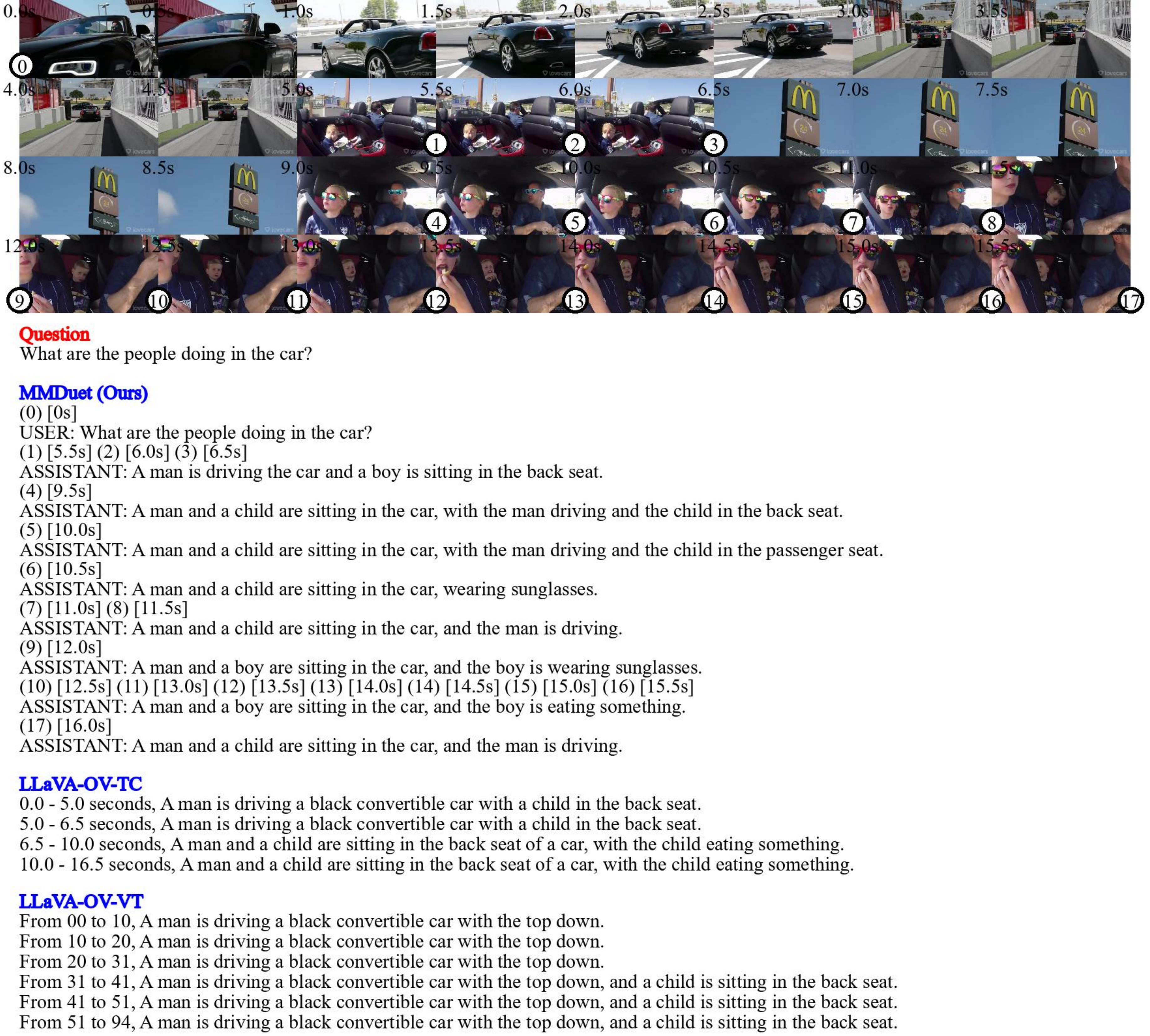}
    \caption{An example of multi-answer grounded video question answering with \modelname, LLaVA-OV-TC and LLaVA-OV-VT.}
    \label{fig:magqa_example1}
\end{figure*}

\begin{figure*}
    \centering
    \includegraphics[width=\textwidth]{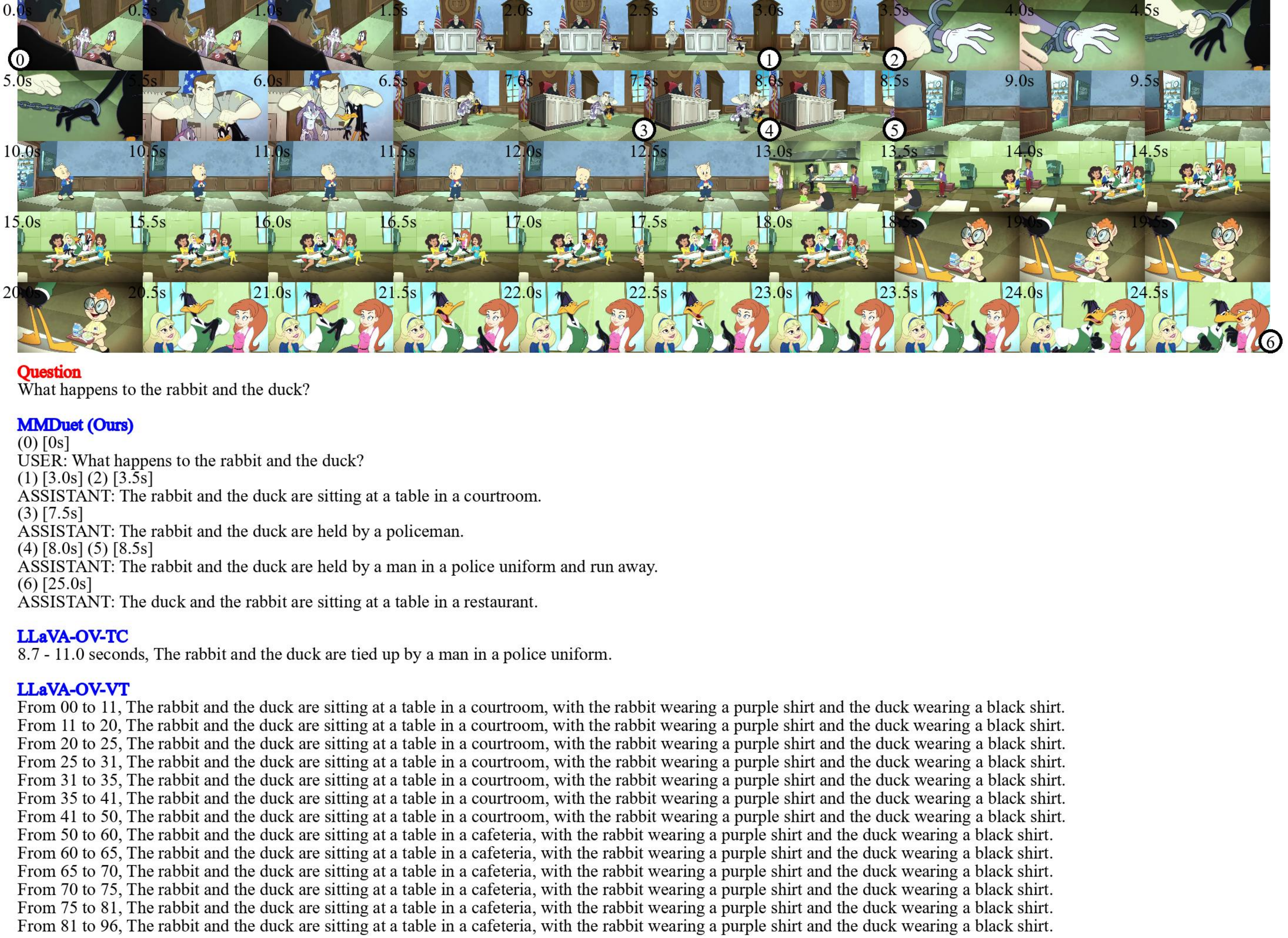}
    \caption{An example of multi-answer grounded video question answering with \modelname, LLaVA-OV-TC and LLaVA-OV-VT.}
    \label{fig:magqa_example2}
\end{figure*}

\begin{figure*}
    \centering
    \includegraphics[width=\textwidth]{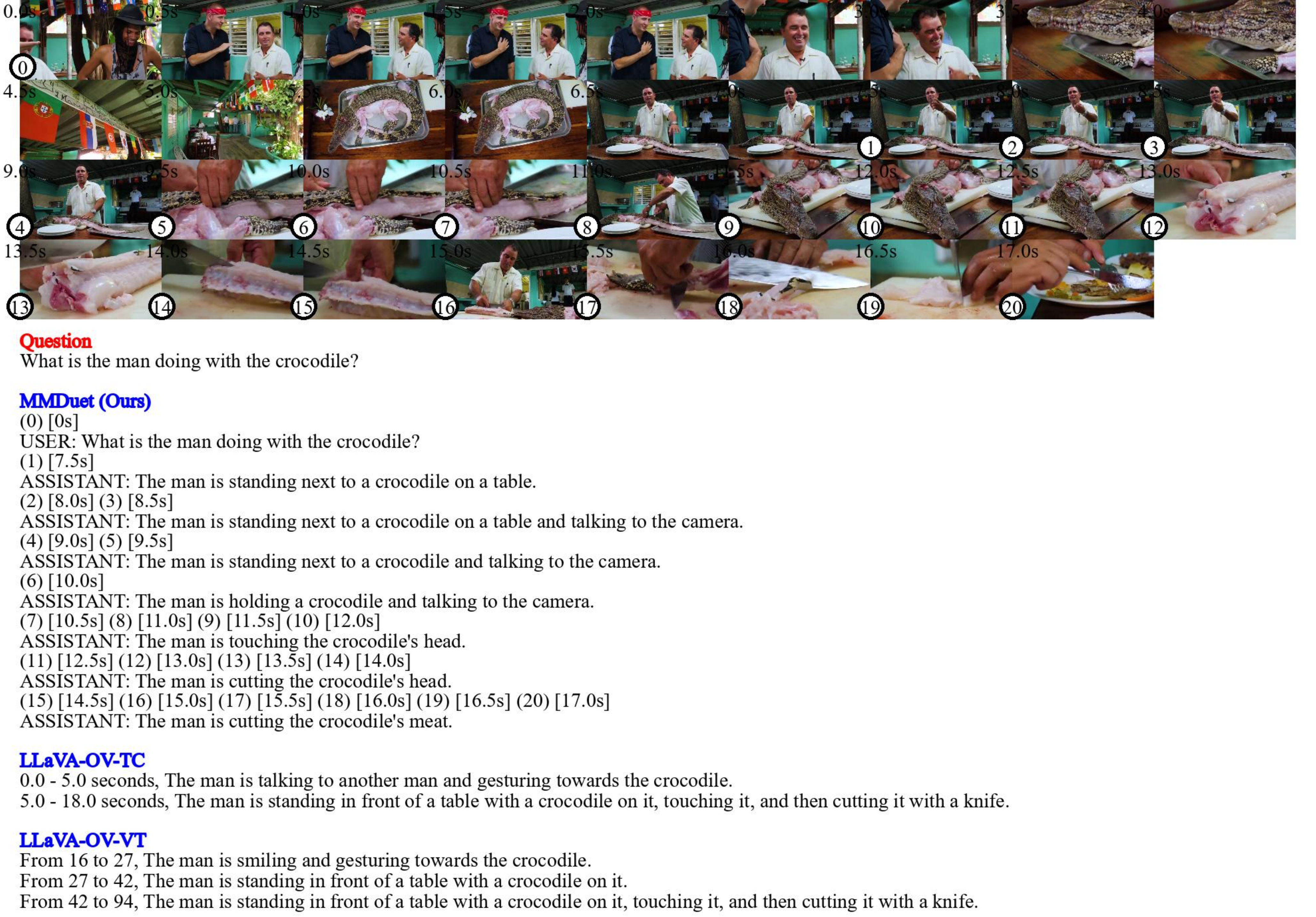}
    \caption{An example of multi-answer grounded video question answering with \modelname, LLaVA-OV-TC and LLaVA-OV-VT.}
    \label{fig:magqa_example3}
\end{figure*}

We list some examples of dense video captioning on videos with several minutes in length and contains many actions in \cref{fig:dvc_example1,fig:dvc_example2,fig:dvc_example3}, and examples of multi-answer grounding video question answering (MAGQA) in \cref{fig:magqa_example1,fig:magqa_example2,fig:magqa_example3}.
For LLaVA-OV-TC and LLaVA-OV-VT, we directly list their generated outputs. For \modelname, we list the numerical order (in round brackets), time (in square brackets) and content (in the second line) for each turn. If a line contains multiple numerical orders and times, this indicates that these turns have the same content, which is shown in the following line. To help readers to identify the position of these turns within the video, we also annotate the numerical order of the turns at the corresponding timestamps in the video stream.

When handling long videos for dense video captioning, baseline models often recall only part of the video or generate repeated content, failing to provide a complete description of all steps in the video. In contrast, \modelname, due to its ability to focus only on a small portion of the video content preceding each generation step and using the ``rm. prev. turns'' trick to avoid interference from previous turns, can provide more accurate and detailed video descriptions.

For the MAGQA task, due to the relatively short video length, baseline models can also locate video segments and answer questions effectively. The advantage of \modelname in this task is its ability to provide answers in a real-time manner. 

\end{document}